\def\year{2021}\relax
\documentclass[letterpaper]{article} 
\usepackage{aaai21}  
\usepackage{times}  
\usepackage{helvet} 
\usepackage{courier}  
\usepackage[hyphens]{url}  
\usepackage{graphicx} 
\usepackage{natbib}  
\usepackage{caption} 
\usepackage{multirow}
\usepackage{bbm}
\usepackage{array}
\usepackage{amsmath}
\usepackage{amssymb}
\usepackage{amsfonts}

\title{Improving Image Captioning by Leveraging Intra- and Inter-layer Global Representation in Transformer Network}
\author {
    Jiayi Ji\textsuperscript{\rm 1},
    Yunpeng Luo\textsuperscript{\rm 1},
    Xiaoshuai Sun\textsuperscript{\rm 1,2}\thanks{Corresponding Author},
    Fuhai Chen\textsuperscript{\rm 1}, \\
    Gen Luo\textsuperscript{\rm 1},
    Yongjian Wu\textsuperscript{\rm 3},
    Yue Gao\textsuperscript{\rm 4},
    Rongrong Ji\textsuperscript{\rm 1,2} \\
}

\affiliations {
    \textsuperscript{\rm 1} Media Analytics and Computing Lab, Department of Artificial Intelligence, \\ School of Informatics, Xiamen University, 361005, China \\
    \textsuperscript{\rm 2} Institute of Artificial Intelligence, Xiamen University \\
    \textsuperscript{\rm 3} Tencent Youtu Lab \qquad\qquad\  
    \textsuperscript{\rm 4} Tsinghua University \\
    jjyxmu@gmail.com,lyricpoem1997@gmail.com,xssun@xmu.edu.cn,chenfuhai3c@163.com, \\ luogen@stu.xmu.edu.cn, littlekenwu@tencent.com, gaoyue@tsinghua.edu.cn, rrji@xmu.edu.cn
}
\begin{document}

\maketitle

\begin{abstract}
Transformer-based architectures have shown great success in image captioning, where object regions are encoded and then attended into the vectorial representations to guide the caption decoding. However, such vectorial representations only contain region-level information without considering the global information reflecting the entire image, which fails to expand the capability of complex multi-modal reasoning in image captioning. In this paper, we introduce a Global Enhanced Transformer (termed GET) to enable the extraction of a more comprehensive global representation, and then adaptively guide the decoder to generate high-quality captions.  In GET, a Global Enhanced Encoder  is designed for the embedding of the global feature,  and a Global Adaptive Decoder are designed for the guidance of the caption generation. The former models intra- and inter-layer global representation by taking advantage of the proposed Global Enhanced Attention and a layer-wise fusion module. The latter  contains a Global Adaptive Controller that can adaptively fuse the global information into the decoder to guide the caption generation.  Extensive experiments on MS COCO dataset demonstrate the superiority of our GET over many state-of-the-arts.
\end{abstract}

\section{INTRODUCTION}

\begin{figure}[t]
	\centering
	\includegraphics[width=0.91\columnwidth]{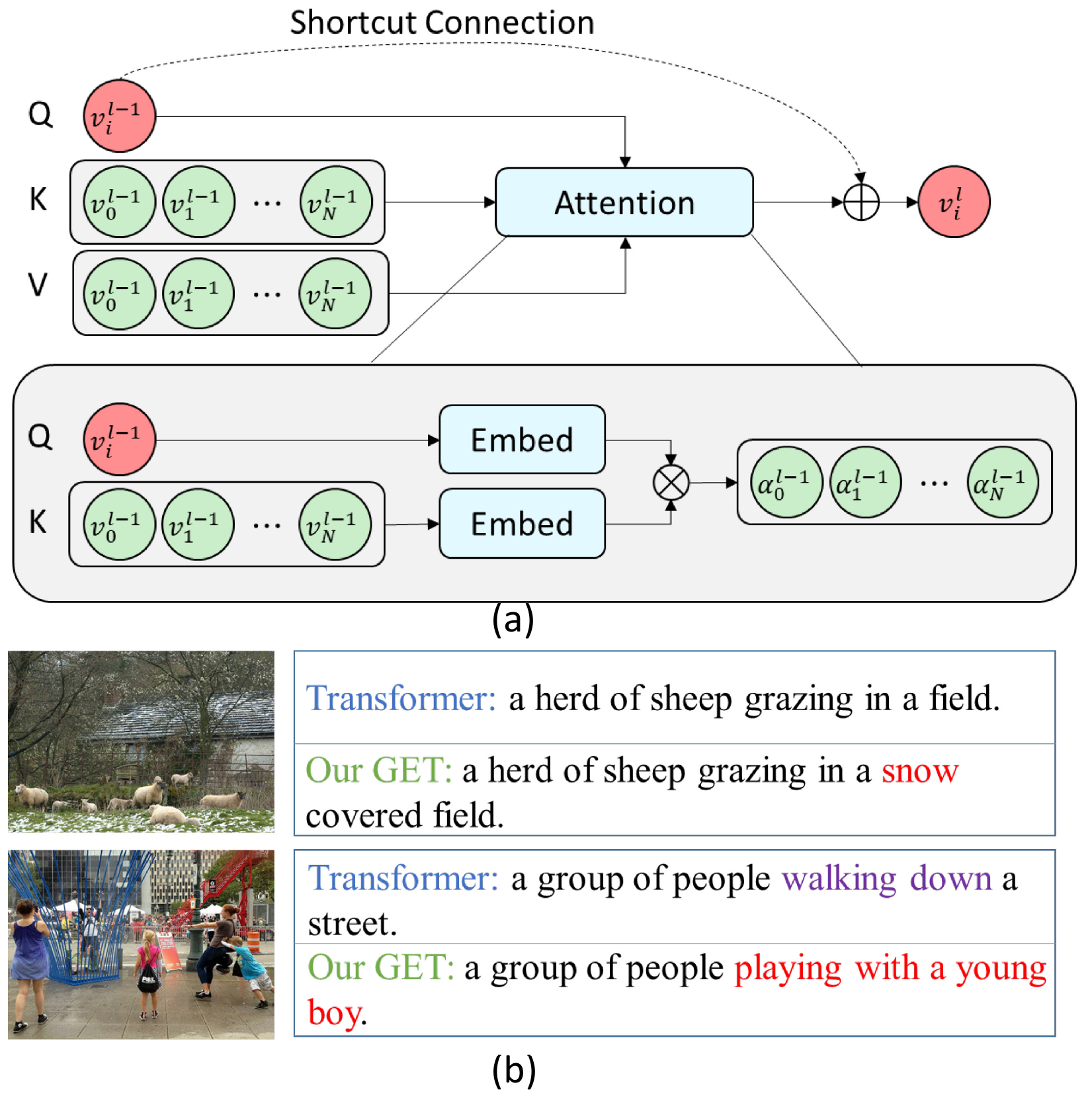} 
	\caption{(a) The self-attention mechanism in the $l$-th layer of a standard Transformer. The vectorial representation $v_i^l$ is region-biased, which only focuses on the region-level information \cite{devlin2018bert,song2020alignment,weng2020gret}. (b) Two key issues of the traditional Transformer-based captioning model that we try to address: object missing (top: missing ``snow'') and false prediction (bottom: predicting ``playing with a boy'' as ``waking down''). }
	\label{fig:figure1}
\end{figure}

Image captioning aims to describe the semantic content  of an image via neural language, which has recently attracted extensive research attention. Inspired by the sequence-to-sequence model for machine translation, most captioning models \cite{vinyals2016show,xu2015show,anderson2018bottom,huang2019attention} mainly adopt a encoder-decoder framework, where an encoder network encodes the input image into a vectorial feature, and a decoder network takes the vectorial feature as input and generates the output caption. Such an encoder-decoder framework is recently well promoted with the development of the Transformer \cite{vaswani2017attention}, where the self-attention is efficiently utilized to capture the correlations among the regions and words \cite{liu2019aligning,huang2019attention,li2019entangled,herdade2019image,cornia2020meshed}.

%
In the Transformer architecture, a set of image regions are encoded and attended into vectorial representations, as shown in Fig. \ref{fig:figure1} (a). These representations are then fused into the decoder to generate the corresponding captions.  
However, as demonstrated by earlier works \cite{devlin2018bert,song2020alignment,weng2020gret}, even though the vectorial representations of these regions are hierarchically calculated by being attended  to all regions in the image, they still ignore the image-level characteristics and are thereby less effective for the decoder \cite{weng2020gret,anderson2018bottom}. 
It causes the problem of object missing when generating descriptions, which is attributed to the limit number of categories in object detectors. As shown in the top of Fig. \ref{fig:figure1} (b), an important concept, \emph{i.e}. ``snow'', is not presented. Besides,  it is more error-prone by focusing on local information while ignoring global guidance, as shown in the bottom of Fig. \ref{fig:figure1} (b), which is attributed to treating each object in isolation,  to lead to a relationship bias.



To improve the caption quality, a natural way is to capture and leverage global representation to guide the selection of attractive objects and their relationships, which is however nontrivial due to two challenges. First, directly extracting a global representation from an image by techniques like pooling might introduce strong contextual noises, which severely  cause semantic ambiguity and damage the representation accuracy. Such damage can be even accumulated for multi-step self-attention in Transformers. Second, the extracted global representation can not be directly used by the Transformer decoder since the need for global guidance varies during the generation of captions. 

\begin{figure*}[t]
	\centering
	\includegraphics[width=0.9\textwidth]{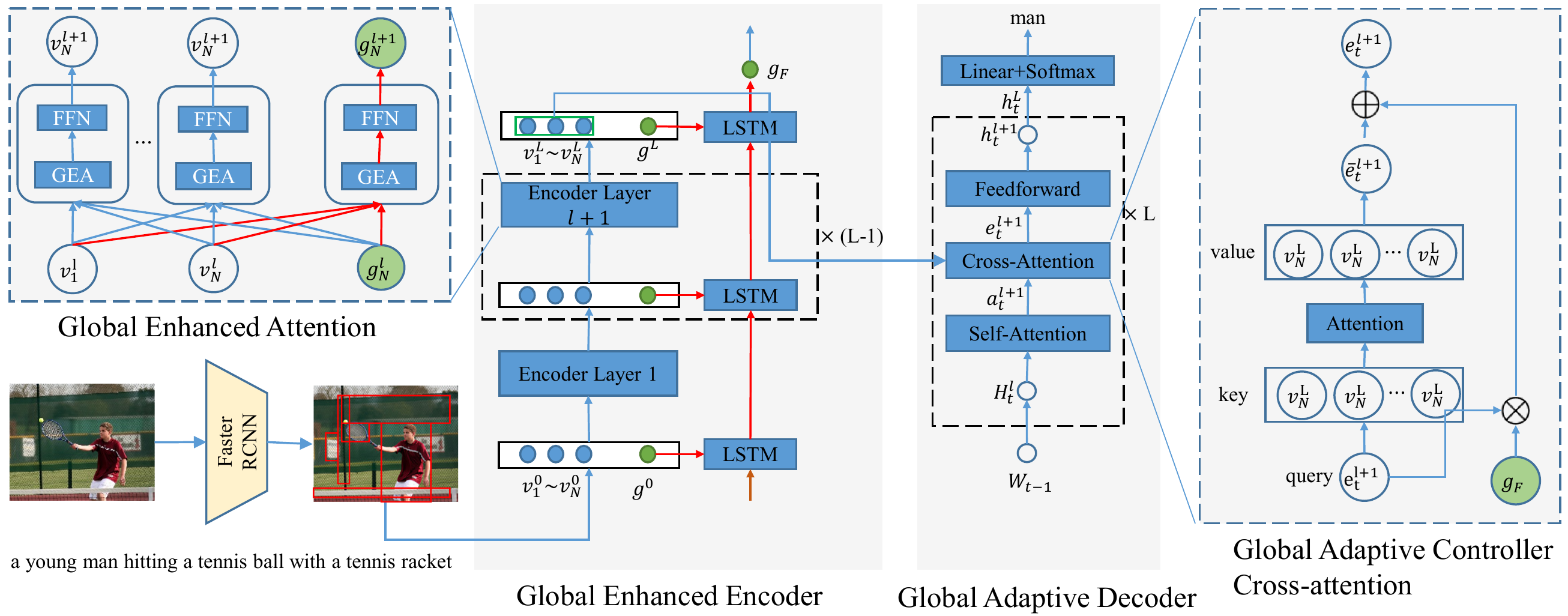} 
	\caption{Overview of our Global Enhanced Transformer Networks (GET) for image captioning. A set of regions are first fed into a global enhanced encoder to extract intra- and inter-layer global information and region-level representation, which are then adaptively fused into the decoder to generate captions. Notice that the Residual Connections, Layer Normalizations, and Embedding Layers are omitted.}
	
	\label{fig:overframework}
\end{figure*}

To solve the above problems, we propose a new Transformer architecture, \emph{i.e.}, Global Enhanced Transformer (termed GET) as shown in Fig. \ref{fig:overframework}. GET captures the global feature via Global Enhanced Attention and utilizes the global feature to guide the caption generation via Gated Adaptive Controller. In GET, we first design a Global Enhanced Encoder to extract intra- and inter-layer global representations. Specifically, we adopt Global Enhanced Attention to aggregate local information from each layer to form intra-layer global representation. After that, the global features are sequentially aggregated among layers via recurrent neural networks, which discard useless information from the previous layers.  Then we adaptively fuse the distilled global representation into the decoder via a Global Adaptive Controller module, which can be implemented by two alternative gating modules to control the fusion, \emph{i.e.,} Gate Adaptive Controller and Multi-Head Adaptive Controller. As the local vectorial representations may be insufficiently comprehensive in detail, GET explores the global parts of images to supplement the local vectorial representation, which could be more comprehensive and instructive for caption generation.

To sum up, our major contributions are itemized below:

\begin{itemize}
	\item We address the issue of object missing and relationship bias by leveraging global represention to provide more comprehensive visual information and play the role of connecting various local parts, which is fundamental in image captioning task.
	\item We devise a unique encoder, termed Global Enhanced Encoder, which enables the Transformer framework to model intra- and inter-layer global information simultaneously, and  propose a novel gating mechanism named Gated Adaptive Controller to provide an adaptive and sophisticated control for the fusion of global information.
	\item Through extensive experiments,  we demonstrate that our Global Enhanced Transformer (GET) model can achieve new state-of-the-art performance on MS COCO dataset.
\end{itemize}

\section{RELATED WORK}

\textbf{Image Captioning.} Inspired by the encoder-decoder architectures in machine translation \cite{bahdanau2014neural,sutskever2014sequence}, most existing image captioning approaches typically adopt the CNN-RNN framework \cite{vinyals2016show,karpathy2015deep}, where a convolution neural network (CNN) \cite{he2016deep,lin2020hrank} is used to encode a given image, which is followed by a recurrent neural network (RNN) \cite{hochreiter1997long} to decode the CNN output into a sentence. Recently, a variety of advanced models \cite{yao2018exploring,yang2019auto,anderson2018bottom,lu2017knowing} have been proposed with attention \cite{xu2015show} and RL-based training objectives \cite{rennie2017self}.  

\noindent \textbf{Transformer-based Image Captioning.} Some recent approaches have explored the use of the Transformer model \cite{vaswani2017attention} in Vision-Language tasks. \cite{huang2019attention} introduced a Transformer-like encoder to encode the regions into the hidden states, which was paired with an LSTM decoder. Recently, \cite{zhu2018captioning,herdade2019image,pan2020x,guo2020normalized,li2019boosted,cornia2020meshed} proposed to replace conventional RNN with the Transformer architecture, achieving new state-of-the-art performance. On the same line, \cite{li2019entangled,liu2019aligning,liu2020exploring} used the Transformer to integrates both visual information and additional semantic concepts given by an external tagger. However, leveraging global information in the Transformer for the image captioning task has never been explicitly explored, which motivates our work in this paper.

\section{PRELIMINARIES} 
The Transformer-based models formulate the calculation of the $t$-th hidden state of decoder as
\begin{equation}
h_t = Decoder(Encoder(I),w_1, \cdots, w_{t-1}), 
\end{equation}
where $w_i$ represents the feature embedding of the $i$-th word. 
The Transformer contains an encoder which consists of a stack of self-attention and feed-forward layers, and a decoder which uses self-attention on textual words and cross-attention over the vectorial representations from the encoder to generate the caption word by word.

We first present a basic form of attention, called ``Scaled Dot-Product Attention'' , which is first proposed as a core component in Transformer \cite{vaswani2017attention}. All intra-modality and cross-modality interactions between word and image-level features are modeled via this basic form of attention.  The attention module operates on some queries $Q$, keys $K$ and values $V$ and generates weighted average vectors $\hat{V}$, which can be formulated as:

\begin{equation}
\hat{V} = \text{ Attention }(Q, K, V)=\operatorname{softmax}\left(\frac{Q K^{T}}{\sqrt{d}}\right) V,
\end{equation}
where $Q$ is a matrix of $n_q$ query vectors, $K$ and $V$ both contain $n_k$ keys and values, all with the same dimensionality, and d is a scaling factor.

To extend the capacity of exploring subspaces, Transformer employs an effective module called multi-head attention, which is defined as
\begin{equation}
{MultiHead}({Q}, {K}, {V})= {Concat}\left({H}_{{1}}, \ldots, {H}_{{h}}\right) {W}^{O},
\end{equation}
\begin{equation}
{H}_{i}= {Attention}\left({Q}{W}_{i}^{Q}, {K W}_{i}^{K}, {V W}_{i}^{V}\right),
\end{equation}
where ${W}_{i}^{Q}, {W}_{i}^{K}, {W}_{i}^{V} \in \mathbb{R}^{\frac{d}{h} \times d}$ are the independent head projection matrices, $i = 1, 2, \cdots , h$, and ${W}^{O}$ denotes the linear transformation.

\section{OUR METHOD}
In this section, we devise our Global Enhanced Transformer (GET) for image captioning. As shown in Fig. \ref{fig:overframework}, the overall architecture follows the encoder-decoder paradigm. First, a global-enhanced encoder maps the original inputs into highly abstract local representations and extracts the intra- and inter-layer global representation. Then the decoder adaptively incorporates the multimodal information simultaneously through the proposed global adaptive controller to generate the caption word by word.

\subsection{Global-enhanced Encoder}
The image is represented as a group of visual features $V = \{v_1, v_2, \cdots, v_N\}$ extracted from a pre-trained object detector as \cite{ren2015faster}, where $N$ is the number of visual regions.  Specifically, the detector is a Faster-RCNN model pre-trained on the Visual Genome dataset \cite{krishnavisualgenome}. We can represent the images as: 
\begin{equation}
g = \frac{1}{N}\sum_{i=1}^{N}v_i.
\end{equation}
Each encoder is a stack of $L$ identical layers, of which each one contains a novel structure, \emph{i.e.}, the global-enhanced self-attention (GEA). To adapt the feature dimensionality to the encoder, the visual features V is first fed into a fully-connected layer, then we get projected features  $V^0 = \{v_1^0, v_2^0, \cdots, v_N^0\}$ and $g^0$. 

\textbf{Global-enhanced attention}. The early methods only feed regions to the encoder to extract the vectorial representation. As shown in \cite{devlin2018bert,song2020alignment,weng2020gret}, even though the vectorial representation of each region is hierarchically calculated by attending to all regions in the image, these vectorial representations only contain local features which focus on the region-level information. 
To capture a comprehensive global representation, both region features $V$ and global feature $g$ are fed into the multi-head self-attention module in each layer. By this way, the local information can be aggregated to form the global representation, through which we can capture the intra-layer global information. Specifically, the output of the $l$-th $(0 \leq l \textless L)$ layer $O^l \in R^{d \times (n+1)}$ is fed into the multi-head self-attention module in the $(l + 1)$-th layer, which is then followed by a residual connection and a layer-normalization:
\begin{equation}
\begin{split}
\overline{V}^{l+1} =& GEA(O^l) \\
=& MultiHead(O^l, O^l, O^l), 
\end{split}
\end{equation}
\begin{equation}
V^{l+1} = LayerNorm(O^l + \overline{V}^{l+1}),
\end{equation}
where $O^0 = (V^0; g^0)$, and the residual connections help avoid the vanishing gradient problem the training phase. Then a final feed-forward neural network is adopted for additional processing of the outputs, which is also followed by a residual connection and a layer normalization step:
\begin{equation}
O^{l+1} = LayerNorm(V^{l+1} + FFN(V^{l+1})),
\end{equation}

As illustrated in \cite{dou2018exploiting,wang2020multi}, the representations in different layers have different meanings. Thus we integrate the global representation from different layers to fuse all the low- and high-level information. Note that such a fusion can also help ease the information flow in the stack \cite{wang2020multi}. A straightforward way is pooling (\emph{e.g.}, average pooling), which however loses layer information. In contrast, we adopt LSTM network \cite{hochreiter1997long} for layer-wise fusion and achieve the final global representation $g_{F}$:
\begin{equation}
h_i = LSTM(g^i,h_{i-1}),  g_{F} = h_L,
\end{equation} 
where the LSTM control the model to forget useless information from previous layers via the forgetting gate, which aggregates the global representation from the first layer to $L$-th layer to obtain inter-layer information.

\subsection{Global Adaptive Decoder}
In the decoding phase, the global representation was adaptively fused into the decoder to guide caption generation.  Similar to the encoder, the decoder consists of $N$ identical layers. We start with the basic layer of the global adaptive decoder, which contains a global adaptive controller (GAC) to decide how much the global contextual information should be considered.

Based on the local representation $V^L$ and global representation $g_F$,  the decoder generates captions for the image word-by-word. Suppose the decoder is generating the $t$-th word in the target sentence. We denote $w_t \in {\mathbb R}^{d \times 1}$as the vector representation of the $t$-th word, which is the sum of word embedding and positional encoding. Therefore, the input matrix representation for time step $t$ is:
\begin{equation}
W_{t-1} = (w_0, w_1, \cdots, w_{t-1}),
\end{equation}
where $w_0$ represents the start of sentence.

For the $(l+1)$-th layer, the inputs  $H^l_t = \{h^l_1, h^l_2 ,\cdots, h^l_t\} \in \mathbb{R}^{d \times t}$ are fed into a multi-head self-attention module:
\begin{equation}
\overline{a}^{l+1}_t = MutiHead(h^l_t, H^l_t, H^l_t).
\end{equation}
Note that $W_{t-1}$ are the inputs of the first layer and $h^0_{t} = w_{t-1}$. Then there is a residual connection around them, which is followed by a layer-normalization step:
\begin{equation}
a^{l+1}_t = LayerNorm(h^l_t + {\overline{a}}^{l+1}_t).
\end{equation}
Subsequently, the output $a^{l+1}_t$ is passed into the other multi-head cross-attention, \emph{e.g.}, GAC to incorporate with features $V$ and $g_F$, which is followed by  a residual connection and a layer-normalization:
\begin{equation}
\overline{e}_t^{l+1} = \textit{GAC}(a^{l+1}_t, V^L, g_F)
\end{equation}
\begin{equation}
e_t^{l+1} = LayerNorm(a^{l+1} + \overline{e}_t^{l+1}),
\end{equation}
where $e_t^{l+1}$ contains multi-model information, which is adaptively refined by the global representation to model a more comprehensive and suitable representation. The detail of GAC is described in the next subsection. Then we feed it into a feed-forward neural network (FFN), which is followed by a residual connection and a layer-normalization to obtain the output:
\begin{equation}
h^{l+1}_t = LayerNorm(e^{l+1}_t + FFN(e_t^{l+1})).
\end{equation}

Finally, the output of layer N is fed into the classifier over vocabulary to predict the next word. Let the predicted caption be $Y_t = \{y_0, y_1, \cdots, y_t\}$, where $y_i \in V$, and V is the vocabulary of the captions. Then the conditional probability distribution of words at time t is $p(y_t|Y_{t-1})$, which  can be calculated by: \begin{equation}
p(y_t|Y_{t-1}) = softmax(W_yh^{L}_t),
\end{equation}
where $W_y \in \mathbb{R}^{|V| \times d}$, and $|V|$ is the number of words in the vocabulary.



\subsection{Global Adaptive Controller Cross-attention}
In the generation process, we design two alternative functions for the global adaptive controller to fuse the global information into decoder according to the contextual signals, \emph{i.e.},  Gate Adaptive Controller (GAC) and Multi-Head Adaptive Controller (MAC).

\textbf{Gate Adaptive Controller Self-Attention.} The demand for global information for each target word is different. Motivated by \cite{lu2017knowing}, we propose a context gating mechanism to control the importance of global information. 
The context gate is determined by the query $a^{l+1}_t$ and the global representation $g_L$:
\begin{equation}
\alpha = sigmoid\big((a^{l+1}_t)^\mathrm{T}g_L\big).
\end{equation}
We then adaptively fuse the global representation to refine the output from multi-head self-attention as below:
\begin{equation}
\hat{e}^{l+1}_t = MultiHead(a^{l+1}_t, V^L, V^L),
\end{equation}
\begin{equation}
\overline{e}_t^{l+1} = {\hat{e}^{l+1}_t} + \alpha * g_L.
\end{equation}

\textbf{Multi-Head Adaptive Controller Self-Attention.} A more sophisticated method is to use  the multi-head attention for fusion, which naturally fuses the local represention and the global representation by taking a weighted sum of region vectors $V^L$ and global vector $g_L$. We set $V_g = (V^L;g_F) \in \mathbb{R}^{(N+1) \times d}$ 
\begin{equation}
\overline{e}_t^{l+1} = MultiHead(a^{l+1}_t, V_g, V_g)
\end{equation}
Noticeably, attentive weights depend solely on the pairwise similarities between visual vectors (\emph{e.g.,} region vectors and global vectors) and the query vector. In such a way, the output can capture suitable global information to refine the original local representation. Besides, the multi-head mechanism allows the model to jointly attend to information from different representation subspaces.

\subsection{Training}
For a given caption $Y_T = \{y_0, \cdots, y_T\}$, the distribution is calculated as the product of the conditional distributions at all time steps:
\begin{equation}
p_l(Y) = \prod_{t=0}^{T}{p({y}_t|Y_{t-1})}.
\end{equation}

The training process consists of two phases: pre-training by supervised learning and fine-tuning by reinforcement learning. Let $\theta$ be the parameters of the model. In pre-training, given a target ground truth sequence $Y^* = \{{y^*_0}, \cdots, {y^*_T}\}$, the objective is to minimize the cross-entropy loss (XE): 
\begin{equation}
L(\theta) = -\sum_{t=0}^{T}log\big(p({y^*_t}|{Y^*_{{t-1}}})\big).
\end{equation}
At the fine-tuning stage, we employ a variant of the self-critical sequence training approach \cite{rennie2017self} on sequences sampled using beam search to directly optimize the  metric, following previous works \cite{rennie2017self,anderson2018bottom}. The final gradient for one sample is calculated as:
\begin{equation}\nabla_{\theta} L(\theta)=-\frac{1}{k} \sum_{i=1}^{k}\bigg(\Big(r\big(\boldsymbol{Y}^{i}\big)-b\Big) \nabla_{\theta} \log p\big(\boldsymbol{Y}^{i}\big)\bigg)\end{equation}

\noindent where $r(\cdot)$ can be any evaluation score metric, and we use the CIDEr-D score as a reward. 
$Y^i = \{y^i_0, \cdots, y^i_T\}$ is the $i$-th sentence in the beam, and $b=\Big(\sum_{i} r\big({Y}^{i}\big)\Big) / k$ is the baseline, computed as the mean of the rewards obtained by the sampled sequences.

\section{EXPERIMENTS}

\begin{table}[]
	\small
	\resizebox{1.00\columnwidth}{!}{ 
		\begin{tabular}{|l|l|l|l|l|l|l|l|l|}
			\hline
			                            &B-1   & B-4  & M    & R    & C    & S    \\ \hline
			SCST                        &   -  & 34.2 & 26.7 & 55.7 & 114.0&  -   \\ \hline
			Up-Down                     & 79.8 & 36.3 & 27.7 & 56.9 & 120.1& 21.4 \\ \hline
			RFNet                       & 79.1 & 36.5 & 27.7 & 57.3 & 121.9& 21.2 \\ \hline
			GCN-LSTM                    & 80.5 & 38.2 & 28.5 & 58.3 & 127.6& 22.0 \\ \hline
			Up-Down+HIP                 &  -   & 38.2 & 28.4 & 58.3 & 127.6& 22.0 \\ \hline
			SGAE                        & 80.8 & 38.4 & 28.4 & 58.6 & 127.8& 22.1 \\ \hline
			ETA                         & \textbf{81.5} & 39.3 & 28.8 & 58.9 & 126.6& 22.7 \\ \hline
			SRT                         & 80.3 & 38.5 & 28.7 & 58.4 & 129.1& 22.4 \\ \hline
			AoANet                      & 80.2 & 38.9 & 29.2 & 58.8 & 129.8& 22.4 \\ \hline
			ORT                         & 80.5 & 38.6 & 28.7 & 58.4 & 128.3& 22.6 \\ \hline
			MMT                         & 80.8 & 39.1 & 29.2 & 58.6 & 131.2& 22.6 \\ \hline
			POS-SCAN                    & 80.2 & 38.0 & 28.5 & -    & 125.9& 22.2 \\ \hline
			CBT                         & -    & 39.0 & 29.1 & \textbf{59.2} & 128.1& \textbf{22.9} \\ \hline
			Ours(w/ GAC)                & 80.8 & 38.8 & 29.0 & 58.6 & 130.5& 22.4 \\ \hline
			Ours(w/ MAC)                & \textbf{81.5} & \textbf{39.5} & \textbf{29.3} & 58.9 & \textbf{131.6}&  22.8 \\ \hline
		\end{tabular}
	}
	\caption{Comparison with the state of the art on the ``Karpathy'' test split, in single-model setting. All values are reported as percentage (\%).}
	\label{tab:offline}
\end{table}

\subsection{Dataset and Implementation Details}
All the experiments are conducted on the most popular benchmark dataset of image captioning, i.e., MS COCO \cite{lin2014microsoft}. The whole MSCOCO dataset contains 123,287 images, which includes 82,783 training images, 40,504 validation images, and 40,775 testing images. Each image is equipped with five ground-truth sentences.  The online evaluation is done on the MS COCO test split, for which ground-truth annotations are not publicly available. In offline testing, we use the Karpathy splits \cite{karpathy2015deep} that have been used extensively for reporting results in previous works. This split contains 113,287 training images, and 5K images respectively for validation and testing.

\begin{figure*}[t]
	\centering
	\includegraphics[width=0.85\textwidth]{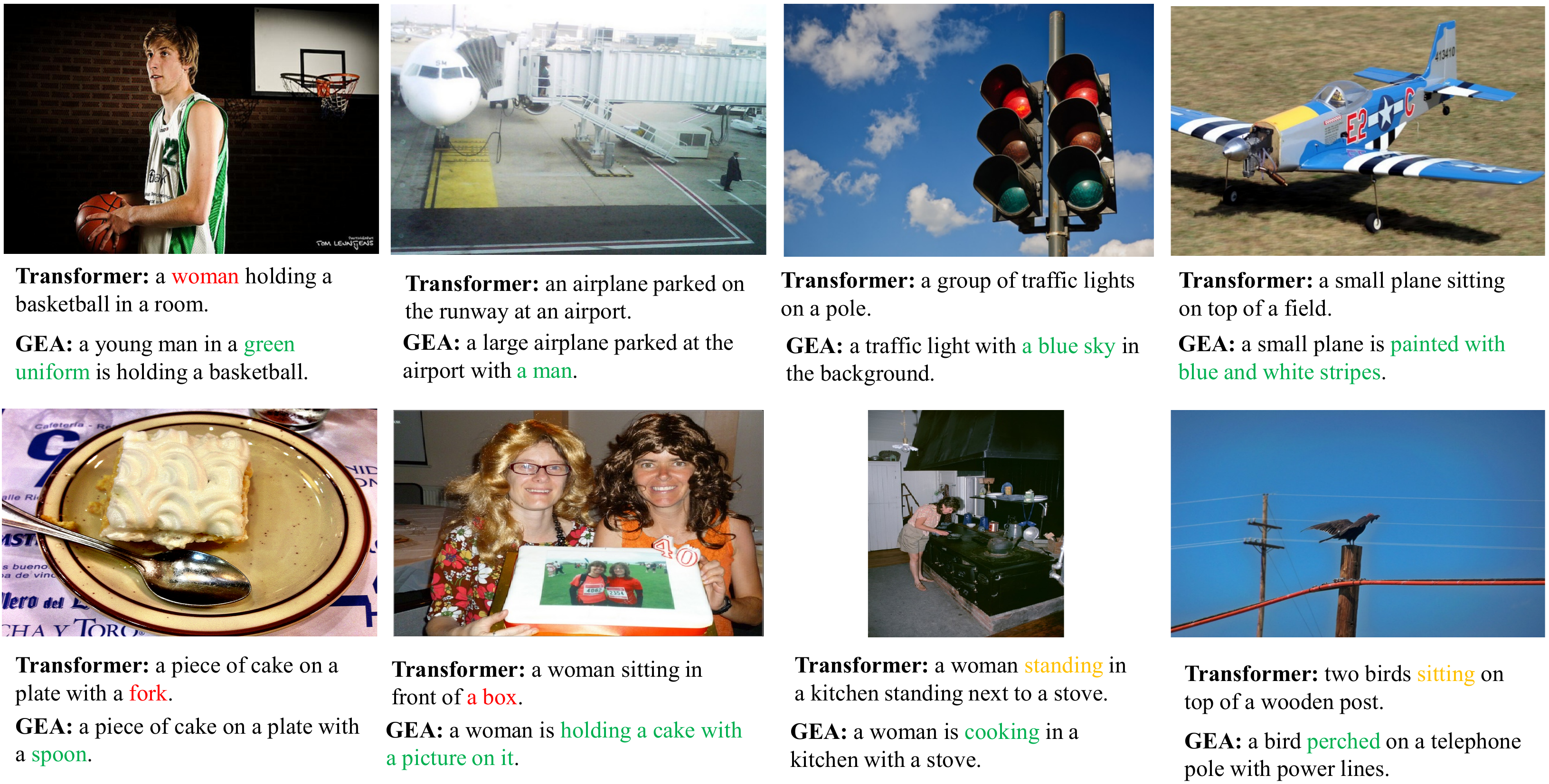} 
	\caption{Examples of captions generated by our approach and the standard Transformer model. Some detailed and accurate words are marked in green, the wrong words are marked in red, and the inaccurate words are marked in yellow. Our method yields more detailed and accurate descriptions.}
	\label{fig:visulization}
	\vspace{-0.2cm}	
\end{figure*}

We use Faster R-CNN \cite{ren2015faster} with ResNet-101 \cite{he2016deep} finetuned on the Visual Genome dataset \cite{krishnavisualgenome} to represent image regions. In our model, we set the dimensionality $d$ of each layer to 512, and the number of heads to 8. We employ dropout with a keep probability of 0.9 after each attention and feed-forward layer. 
Pre-training with XE is done following the learning rate scheduling strategy with a warmup equal to 10, 000 iterations. Then, during CIDEr-D optimization, we use a fixed learning rate of $5 \times 10^{-6}$. We train all models using the Adam optimizer \cite{kingma2014adam}, a batch size of 50, and a beam size equal to 5.
At the inference stage, we adopt the beam search strategy and set the beam size as 3. Five evaluation metrics, \emph{i.e.}, BLEU \cite{papineni2002bleu}, METEOR \cite{banerjee2005meteor}, ROUGE-L \cite{lin2004rouge}, CIDEr \cite{vedantam2015cider}, and SPICE \cite{anderson2016spice}, are simultaneously utilized to evaluate our model.

\begin{table}[]
	\small
	\resizebox{1.00\columnwidth}{!}{ 
		\begin{tabular}{|l|l|l|l|l|l|l|}
			\hline
			Model        & B-1   & B-4   & M    & R    & C     & S    \\ \hline
			\multicolumn{7}{|c|}{Ensemble/Fusion of 2 models}       \\ \hline
			GCN-LSTM     & 80.9 & 38.3 & 28.6 & 58.5 & 128.7 & 22.1 \\ \hline
			SGAE         & 81.0 & 39.0 & 28.4 & 58.9 & 129.1 & 22.2 \\ \hline
			ETA          & 81.5 & 39.9 & 28.9 & 59.0 & 127.6 & 22.6 \\ \hline
			GCN-LSTM+HIP & -    & 39.1 & 28.9 & 59.2 & 130.6 & 22.3 \\ \hline
			MMT          & 81.6 & 39.8 & 29.5 & 59.2 & 133.2 & 23.1 \\ \hline
			Ours         & \textbf{81.9} & \textbf{40.3} & \textbf{29.6} & \textbf{59.4} & \textbf{133.5} & \textbf{23.3} \\ \hline
			\multicolumn{7}{|c|}{Ensemble/Fusion of 4 models}       \\ \hline
			SCST         & -    & 35.4 & 27.1 & 56.6 & 117.5 & -    \\ \hline
			RFNet        & 80.4 & 37.9 & 28.3 & 58.3 & 125.7 & 21.7 \\ \hline
			AoANet       & 81.6 & 40.2 & 29.3 & 59.4 & 132.0 & 22.8 \\ \hline
			MMT          & 82.0 & 40.5 & 29.7 & 59.5 & 134.5 & 23.5 \\ \hline
			Ours         & \textbf{82.1} & \textbf{40.6} & \textbf{29.8} & \textbf{59.6} & \textbf{135.1} & \textbf{23.8} \\ \hline
		\end{tabular}
	}
	\caption{Comparison with the state of the art on the ``Karpathy'' test split, using ensemble technique, where B-N, M, R, C and S are short for BLEU-N, METEOR, ROUGE-L, CIDEr and SPICE scores. All values are reported as percentage (\%).}
	\label{tab:fusion}
\end{table}

\begin{table*}[]
	\centering
	\small
	\begin{tabular}{|l|l|l|l|l|l|l|l|l|l|l|l|l|l|l|}
		\hline
		Model    & \multicolumn{2}{l|}{BLEU-1}   & \multicolumn{2}{l|}{BLEU-2}   & \multicolumn{2}{l|}{BLEU-3}   & \multicolumn{2}{l|}{BLEU-4}   & \multicolumn{2}{l|}{METEOR}   & \multicolumn{2}{l|}{ROUGE-L}  & \multicolumn{2}{l|}{CIDEr-D}    \\ \hline
		Metric   & c5            & c40           & c5            & c40           & c5            & c40           & c5            & c40           & c5            & c40           & c5            & c40           & c5             & c40            \\ \hline
		SCST     & 78.1          & 93.7          & 61.9          & 86.0          & 47.0          & 75.9          & 35.2          & 64.5          & 27.0          & 35.5          & 56.3          & 70.7          & 114.7          & 116.0          \\ \hline
		LSTM-A   & 78.7          & 93.7          & 62.7          & 86.7          & 47.6          & 76.5          & 35.6          & 65.2          & 27.0          & 35.4          & 56.4          & 70.5          & 116.0          & 118.0          \\ \hline
		Up-Down  & 80.2          & 95.2          & 64.1          & 88.8          & 49.1          & 79.4          & 36.9          & 68.5          & 27.6          & 36.7          & 57.1          & 72.4          & 117.9          & 120.5          \\ \hline
		RF-Net   & 80.4          & 95.0          & 64.9          & 89.3          & 50.1          & 80.1          & 38.0          & 69.2          & 28.2          & 37.2          & 58.2          & 73.1          & 122.9          & 125.1          \\ \hline
		GCN-LSTM & -             & -             & 65.5          & 89.3          & 50.8          & 80.3          & 38.7          & 69.7          & 28.5          & 37.6          & 58.5          & 73.4          & 125.3          & 126.5          \\ \hline
		SGAE     & 81.0          & 95.3          & 65.6          & 89.5          & 50.7          & 80.4          & 38.5          & 69.7          & 28.2          & 37.2          & 58.6          & 73.6          & 123.8          & 126.5          \\ \hline
		AoANet   & 81.0          & 95.0          & 65.8          & 89.6          & 51.4          & 81.3          & 39.4          & 71.2          & 29.1          & 38.5          & 58.9          & 74.5 & 126.9          & 129.6          \\ \hline
		ETA      & 81.2          & 95.0          & 65.5          & 89.0          & 50.9          & 80.4          & 38.9          & 70.2          & 28.6          & 38.0          & 58.6          & 73.9          & 122.1          & 124.4          \\ \hline
		MMT      & \textbf{81.6}          & 96.0          & 66.4          & 90.8          & 51.8          & 82.7          & \textbf{39.7}          & 72.8          & \textbf{29.4}          & \textbf{39.0}          & \textbf{59.2}          & \textbf{74.8}          & 129.3          & 132.1          \\ \hline
		Ours     & \textbf{81.6}      & \textbf{96.1} & \textbf{66.5}  & \textbf{90.9} & \textbf{51.9} & \textbf{82.8} & \textbf{39.7} & \textbf{72.9} & \textbf{29.4} & 38.8 & 59.1 & 74.4          & \textbf{130.3} & \textbf{132.5} \\ \hline
	\end{tabular}
	\caption{MS COCO Online Evaluation. All values are reported as percentage (\%), with the highest value of each entry highlighted in boldface.}
	\label{tab:online}
\end{table*}

\subsection{Performance Comparison}
\begin{figure*}[t]
	\centering
	\includegraphics[width=0.9\textwidth]{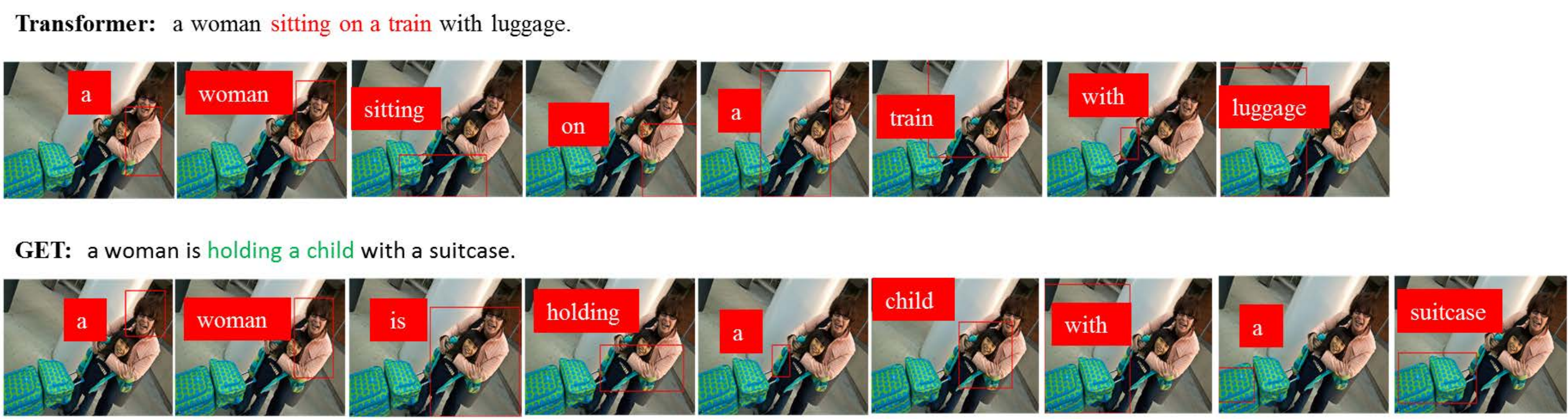} 
	\caption{The visualization of attended image regions along with the caption generation process for plain Transformer and the proposed GET. At the decoding step for each word, we outline the image region with the maximum output attribution in red.}
	\label{fig:visulization2}
	
\end{figure*}

\textbf{Offline Evaluation.} Tab. \ref{tab:offline} and Tab. \ref{tab:fusion} show the performance comparisons between the state-of-the-art models and our proposed approach on the offline COCO Karpathy test split. We show the performances for both the single model version and the ensemble version. The baseline models we compared include SCST \cite{rennie2017self}, LSTM-A \cite{yao2017boosting}, Up-Down \cite{anderson2018bottom}, RFNet \cite{ke2019reflective}, GCN-LSTM \cite{yao2018exploring}, SGAE \cite{yang2019auto}, AoANet \cite{huang2019attention} ORT \cite{herdade2019image}, ETA \cite{li2019entangled}, MMT \cite{cornia2020meshed}, SRT \cite{wang2020show}, POS-SCAN \cite{zhou2020more} and CBT \cite{wang2020compare}. We present the results of the proposed GET with two different global adaptive controllers (\emph{e.g.,} GAC and MAC). For clarity, the symbol ``ours'' only represents the latter one in the following section.

\textbf{Single model.} In Tab.\,\ref{tab:offline}, we report the performance of our method in comparison with the aforementioned state-of-the-art methods, using captions predicted from a single model and optimization on the CIDEr score. Our method surpasses all other approaches in terms of BLEU-4, METEOR and CIDEr, and achieves competitive performance on SPICE and ROUGE-L compared to the SOTA. In particular, it advances the current state of the art on CIDEr by 0.4\%.

\textbf{Ensemble model.} Following the common practice \cite{rennie2017self,huang2019attention} of building an ensemble of models, we also report the performances of our approach when averaging the output probability distributions of multiple and independently trained instances of our model. In Tab. \ref{tab:fusion}, we use ensembles of two and four models, trained from different random seeds. Noticeably, when using four models, our approach achieves the best performance according to all metrics, with an increase of 0.6 CIDEr points with respect to the current state of the art \cite{cornia2020meshed}.

\textbf{Online Evaluation.} Finally, we also report the performance of our method on the online COCO test server. In this case, we use the ensemble of four models previously described, trained on the “Karpathy” training split. Results are reported in Tab. \ref{tab:online}, in comparison with the top-performing approaches on the leaderboard. For fairness of comparison, they also used an ensemble configuration. As can be seen, our method surpasses the current state of the art on most of the metrics, achieving an improvement of 1.0 CIDEr points with respect to the best performer.

\textbf{Qualitative Analysis.} Fig. \ref{fig:overframework} shows several image captioning results of the plain Transformer and our GET.
Generally, compared with the captions of the plain Transformer which are somewhat relevant to image content and logically correct, our GET produces more accurate and descriptive sentences by exploiting intra- and inter-modal interactions. For example, our GET generates the phrase of ``a green uniform'' and ``a man'', while they are missing from the plain Transformer. Besides, our GET generates more precise phrases, such as ``holding a cake with a picture on it'' and ``cooking''. These also confirm the advantage of capturing and leveraging the intra- and inter-layer global representation in the Transformer architectures.



\subsection{Experimental Analysis}

\textbf{Ablation Study.} To validate the effectiveness of our proposed modules, we conduct ablation studies by comparing different variants of the GET. 

Firstly, we investigate the impact of the number of the encoding and decoding layers on captioning performance for the GET. As shown in Tab. \ref{tab:ablation_layer}, varying the number of layers, we observe a slight decrease in performance when increasing the number of layers. Following this finding, all subsequent experiments uses three layers.

\begin{table}[]
	
	\begin{tabular}{|l|l|l|l|l|}
		\hline
		Layer & BLUE-4        & METEOR        & ROUGE-L       & CIDEr          \\ \hline
		2     & 38.2          & 28.9          & 58.3          & 129.7          \\ \hline
		3     & \textbf{39.5} & \textbf{29.2} & \textbf{58.9} & \textbf{131.6} \\ \hline
		4     & 39.2          & 29.2          & 58.6          & 130.7          \\ \hline
		5     & 39.0          & 28.9          & 58.5          & 130.3          \\ \hline
		6     & 39.0          & 29.0          & 58.5          & 130.3          \\ \hline
	\end{tabular}
	\caption{Ablation on the number of encoding and decoding layers. All values are reported as percentage (\%).}
	\label{tab:ablation_layer}
\end{table}

Then, we investigate the impact of all the proposed modules in both encoder and decoder. We choose the plain Transformer as the baseline, which is shown in the third line in Tab. \ref{tab:ablation}. Then we extend the baseline model by adopting the GEA module, which slightly improves the performance. The results indicate that the GEA module can also improve the region level presentation via aggregate information from global representation. Then we investigate the impact of different global representations. As shown in the $5$-th line and $6$-th line, the performance improvements validate the effectiveness of GEA to obtain better presentation than the original presentation $g_0$ via aggregating the intra-layer information. Then we exploit different strategies to fuse the inter-layer information, and the LSTM network obtains the best performance, which basically validates the effectiveness of such layer-wise global representation.  Both the GAC and MAC  gain expected performance, which further indicates the effectiveness of our intra- and inter-layer global representation. And MAC is the better one, which shows that the Multi-Head mechanism works better at feature fusion for its ability of complex relationship modeling.

\begin{table}[]
	\center
	\small
	\resizebox{1.00\columnwidth}{!}{ 
		\begin{tabular}{|c|c|c|c|c|c|c|c|}
			\hline
			\multicolumn{2}{|c|}{encoder} & \multicolumn{1}{l|}{\multirow{2}{*}{decoder}} & \multirow{2}{*}{B-4} & \multirow{2}{*}{M} & \multirow{2}{*}{R} & \multirow{2}{*}{C} \\ \cline{1-2}
			intra-layer   & inter-layer  & \multicolumn{1}{l|}{}                         &                     &                    &                    &                    \\ \hline
			-             & -            & -                                             & 37.9                & 28.0               & 57.9               & 128.1              \\ \hline
			GEA           & -            & -                                             & 38.1                & 28.1               & 58.2               & 128.3              \\ \hline
			$g_0$         & -            & MAC                                           & 38.2                & 28.3               & 58.0               & 128.6              \\ \hline
			GEA           & -            & MAC                                           & 38.4                & 28.3               & 58.2               & 128.9              \\ \hline
			GEA           & average      & MAC                                           & 38.5                & 28.7               & 58.1               & 129.4              \\ \hline
			GEA           & attention    & MAC                                           & 38.7                & 29.0               & 58.2               & 129.8              \\ \hline
			GEA           & LSTM         & MAC                                           & \textbf{39.5}       & \textbf{29.2}      & \textbf{58.9}      & \textbf{131.6}     \\ \hline
			GEA           & LSTM         & GAC                                           & 38.8                & 29.0               & 58.6               & 130.5              \\ \hline
		\end{tabular}
	}
	\caption{Ablation on different variants of the Transformer. All values are reported as percentage (\%).}
	\label{tab:ablation}
\end{table}

\textbf{Attention Visualization.} In order to better qualitatively evaluate the generated results with GET, we visualize the evolutions of the contribution of detected regions to the model output along with the caption generation processes for plain Transformer and the proposed GET in Fig. \ref{fig:visulization2}. The contribution of one region with respect to the output is given by complex non-linear dependencies, which cannot be extracted easily. Therefore, we employ the Integrated Gradients approach \cite{sundararajan2017axiomatic}, which approximates the integral of gradients with respect to the given input via a summation. Results presented in Fig. \ref{fig:visulization2} show that our approach can help to ground the correct image regions to words by exploring the proposed global representation. 


\section{CONCLUSION}
In this paper, we present Global Enhanced Transformer (GET) for image captioning. GET addresses the problem of traditional Transformer-based architectures on the ignorance of global contextual information that limits the capability of reasoning in image captioning. Our model incorporates the Global Enhanced Encoder which captures both intra- and inter-layer global representation to provide more comprehensive visual information and play the role of connecting various local parts, and the Global Adaptive Decoder which adaptively fuses the global information into the decoder to guide caption generation. We show the superior performance of the proposed GET both quantitatively and qualitatively on the MS COCO datasets.

\section{ Acknowledgments}
This work is supported by the National Science Fund for Distinguished Young (No.62025603), the National Natural Science Foundation of China (No.U1705262, No. 62072386, No. 62072387, No. 62072389, No. 62002305,  
No.61772443, No.61802324 and No.61702136) and and Guangdong Basic and Applied Basic Research Foundation (No.2019B1515120049).

\bibliography{GETv_6.1}

\end{document}